\documentclass[conference]{IEEEtran}
\IEEEoverridecommandlockouts
\usepackage{cite}
\usepackage{amsmath,amssymb,amsfonts}
\usepackage{algorithmic}
\usepackage{graphicx}
\usepackage{textcomp}
\usepackage{xcolor}
\def\BibTeX{{\rm B\kern-.05em{\sc i\kern-.025em b}\kern-.08em
    T\kern-.1667em\lower.7ex\hbox{E}\kern-.125emX}}
    
\usepackage{algorithm}
\usepackage{float}
\usepackage{subcaption} 
\usepackage{tabularx}
\usepackage{authblk}
\usepackage{hyperref}
\usepackage{amsthm,bm}

\usepackage{color, colortbl}
\definecolor{LightCyan}{rgb}{0.88,1,1}
\definecolor{Gray}{gray}{0.9}
\usepackage{slashbox}

\begin{document}

\title{Unmasking DeepFakes with simple Features}

\author{Ricard Durall$^{1,2,3}$ \qquad
Margret Keuper$^{4}$ \qquad
Franz-Josef Pfreundt$^{1}$ \qquad
Janis Keuper$^{1,5}$\\
$^1$Fraunhofer ITWM, Germany\\
$^2$IWR, University of Heidelberg, Germany\\
$^3$Fraunhofer Center Machine Learning, Germany\\
$^4$Data and Web Science Group, University Mannheim, Germany\\
$^5$Institute for Machine Learning and Analytics, Offenburg University, Germany\\
}

\maketitle

\begin{abstract}

Deep generative models have recently achieved impressive results for many
	real-world applications, successfully generating high-resolution and
	diverse samples from complex data sets. Due to this improvement, fake
	digital contents have proliferated growing concern and spreading
	distrust in image content, leading to an urgent need for automated ways
	to detect these AI-generated fake images.

Despite the fact that many face editing algorithms seem to produce realistic
human faces, upon closer examination, they do exhibit artifacts in certain
domains which are often hidden to the naked eye. In this work, we present a
	simple way to detect such fake face images - so-called \textit{DeepFakes}.  
Our method is based on a classical frequency domain analysis followed by a basic classifier.
Compared to previous systems, which need to be fed with large amounts of labeled data, our
approach showed very good results using only a few annotated training samples 
	and even achieved good accuracies in fully unsupervised scenarios.
For the evaluation on high resolution face images, we combined several public data sets 
	of real and fake faces into a new benchmark: \textit{Faces-HQ}. Given such high-resolution images, our approach 
	reaches a perfect classification accuracy of 100\% when it is trained
	on as little as 20 annotated samples. In a second experiment, in the
	evaluation of the medium-resolution images of the \textit{CelebA}
	data set, our method achieves 100\% accuracy supervised and 96\%  in an unsupervised
	setting. Finally, evaluating a low-resolution video sequences of the
	\textit{FaceForensics++} data set, our method achieves 90\% accuracy detecting manipulated videos.\\

	Source Code:
\textit{\normalfont{\url{https://github.com/cc-hpc-itwm/DeepFakeDetection}}}\\

\end{abstract}

\begin{IEEEkeywords}
GAN images, DeepFake, Image forensic, Forgery detection
\end{IEEEkeywords}

\section{Introduction}

Over the last years, the increasing sophistication of smartphones and the
growth of social networks have led to a gigantic amount of new digital object
contents. This tremendous use of digital images has been followed by a rise of
techniques to alter image contents. Until recently, such techniques were beyond
the reach of most users since they were dull and time-consuming and they
required a high domain expertise on computer vision. Nevertheless, thanks to
the recent advances of machine learning and the accessibility to large-volume
training data, those limitations have gradually faded away. As a consequence,
the time for fabrication and manipulation of digital contents has significantly
decreased, allowing even amateur users the modification of
contents at their will.

In particular, deep generative models have lately been extensively used to
produce artificial images with realistic appearance. Theses models are based on deep neural networks
which are able to approximate the true data distribution
of a given training set. Hence, one can sample from the learned distribution and add 
variations. Two of the most commonly used and efficient approaches are
Variational Autoencoders (VAE) \cite{kingma2013auto} and Generative Adversarial
Networks (GAN)\cite{goodfellow2014generative}. Especially GAN approaches have lately been 
pushing the limits of state-of-the-art results, improving the resolution and
quality of images produced
\cite{karras2017progressive,brock2018large,karras2019style}. As a result, deep
generative models are opening the door to a new vein of AI-based fake image
generation leading to a fast dissemination of high quality tampered image
content. While significant developments have been made for image forgery
detection, it still remains a hard task since most current methods rely on deep learning
approaches, which require large amounts of labeled training data. 

\begin{figure}[t]
\centering
   \includegraphics[width=\linewidth]{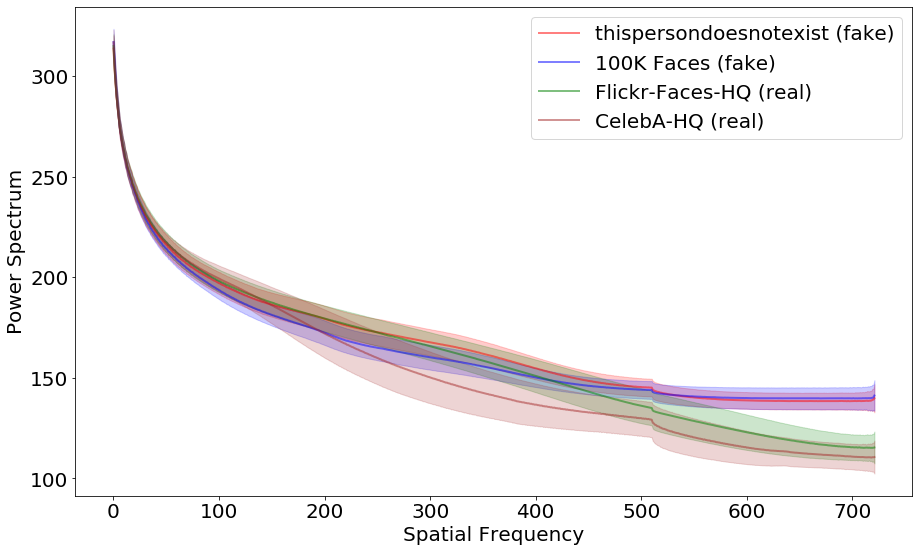}
   \caption{1D power spectrum statistics from each sub-data set from Faces-HQ. The higher the frequency, the bigger is the difference between real or fake data.}
\label{fig:1000}
\end{figure}

\begin{figure*}[t]
\centering
   \includegraphics[width=\linewidth]{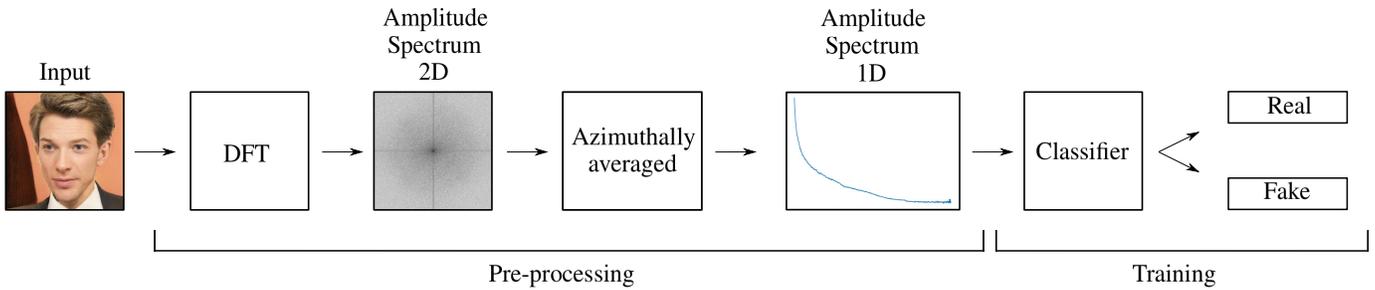}
   \caption{Overview of the processing pipeline of our approach. It contains two main blocks, a feature extraction block using DFT and a training block, where a classifier uses the new transformed features to determine whether the face is real or not. Notice that input images are transformed to grey-scale before DFT.}
\label{fig:pipeline}
\end{figure*}

In this paper, we address the problem of detecting these artificial image
contents, more specifically, fake faces. In order to determine the nature of
these pictures, we introduce a new machine learning based method. Our approach
relies on a classical frequency analysis of the images that reveals different behaviors
at high frequencies. Fig. \ref{fig:1000} shows how different a certain range of
frequency components behave when the images have been artificially generated.

Our method detects such artifacts by analyzing the frequency domain followed by a
simple supervised or unsupervised classifier. Notice that this suggested
pipeline does not involve nor requires vast quantities of data, which is a very
convenient property for those scenarios that suffer from data scarcity.
In addition, we introduce a new data set \textit{Faces-HQ}, which we used to complement the \textit{CelebA} data set and 
\textit{FaceForensics++} data set \cite{roessler2019faceforensicspp}, for our experimental evaluation.\\

Overall, our contributions are summarized as follows:

\begin{itemize}
\item We introduce a novel classification pipeline for artificial face detection based on a frequency domain analysis.

\item We provide a public data set (\textit{Faces-HQ}) of high quality images containing real and fake faces from a set of different public databases.

\item We demonstrate how we successfully learn to detect forgery: extensive experiments on high and medium-resolution images of the \textit{Faces-HQ} and \textit{CelebA} data sets showed 100\% accuracy. Additionally, the evaluation of the \textit{FaceForensics++} data set with low-resolution videos reached 91\% accuracy.

\end{itemize}

\section{Related Work}

In this section, we briefly review the related seminal work on high-resolution artificial images forensics and
deepfake images, as well as forgery
detection. In particular, we focus our attention on spotting tampered images
generated by GAN-based methods.

Traditional image forensics methods can be classified according to the image features that they
target, such as local  noise estimation \cite{pan2012exposing}, pattern
analysis \cite{ferrara2012image}, illumination modeling \cite{de2013exposing} and
steganalysis feature classification \cite{cozzolino2014image}. However, with
the deep learning breakthrough, the computer vision community has radically
steered towards neural networks techniques. For example,
\cite{zhou2017two,cozzolino2019noiseprint} are recent works based on
Convolutional Neural Networks (CNN). These CNN-based approaches also aim to
capture the aforementioned image features, but in an inexplicit way. 

In 2014, \textit{Goodfellow et al.} introduced an adversarial framework (GAN)
which marked a milestone in generative models. In particular, the image
generation has been improved significantly, leading to a striking progress on
artificial faces \cite{brundage2018malicious} among others. As a consequence,
new image and video manipulation techniques known as DeepFake have emerged and
established themselves online over the last few months. This occurrence of
events on digital image forensics has been drawing an ever increasing attention
trying to detect GAN generated images or videos.

The lack of eye blinking \cite{li2018ictu} is one drawback observed, when the videos are artificially created. This is due to the
scarcity of training images including photographs with the subject's eyes
closed. Nevertheless, this detection can be circumvented by adding images with
closed eyes in training. Finding unnatural head poses, is also an extend
technique \cite{yang2019exposing}, in order to detect tampered digital
contents. On the other hand, the works
\cite{mccloskey2018detecting,li2018detection} analyze the color-space features
from GAN generated images  and real images, and use the disparity to classify
them.

Other approaches \cite{marra2018detection,afchar2018mesonet,cispa2965}, rather than
leveraging explicit lacks or failures, rely on CNNs to distinguish GAN's output
from real images. In the same vein,\cite{hsu2018learning} introduces deep
forgery discriminator with a contrastive loss function and
\cite{guera2018deepfake} incorporates temporal domain information by employing
Recurrent Neural Networks (RNNs) upon CNNs. While deep learning methods show
promising performance, a key concern is that all these methods can be easily
learnt by the GAN. In particular, by incorporating them  in to the GAN's
discriminator, the generator can be fine-tuned to learn a countermeasure for
any differentiable forensic.\\

\section{Method}

In the following section, we describe our approach in detail.
Figure \ref{fig:pipeline}) gives an overview of the processing steps.\\

\subsection{Frequency Domain Analysis}

Frequency domain analysis is of utmost importance in signal processing theory and
applications. In particular in the computer vision domain, the repetitive nature
or the frequency characteristics of images can be analyzed on a space defined
by Fourier transform. Such transformation consists in a spectral decomposition
of the input data indicating how the signal's energy is distributed over a range of
frequencies. Methods based on frequency domain analysis have shown wide
applications in image processing, such as image analysis, image filtering,
image reconstruction and image compression.\\

\subsubsection{Discrete Fourier Transform}
The Discrete Fourier Transform (DFT) is a mathematical technique to decompose a
discrete signal into sinusoidal components of various frequencies ranging from 0 
(i.e., constant frequency, corresponding to the image mean value) up to the maximum representable frequency, given the spatial resolution. 
It is the discrete analogon of the continuous Fourier Transform for signals sampled on equidistant points. 
For 2-dimensional data of size $M\times N$, it can be computed as

\begin{align}
\begin{split}
	X_{k,\ell} = \sum\limits_{n = 0 }^{N-1} \sum\limits_{0}^{M-1}{x_{n,m}} \cdot e^{-\frac{i2\pi}{N}kn}\cdot e^{-\frac{i2\pi}{M}\ell m}.
\end{split}
\end{align}

\begin{figure}[t]
\centering
   \includegraphics[width=\linewidth]{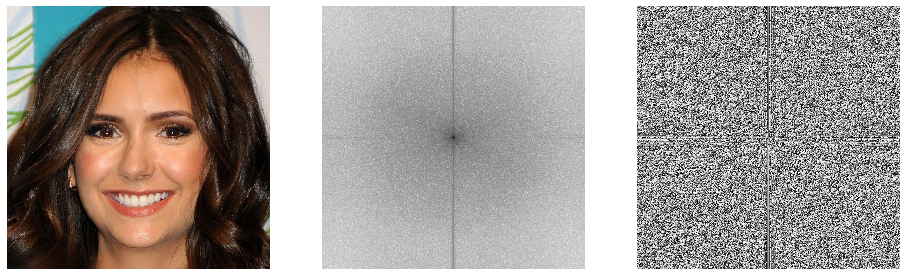}
   \caption[]{Example of a DFT applied to a sample. (Left) Input image \footnotemark. (Center) Power Spectrum. (Right) Phase Spectrum. }
\label{fig:dft}
\end{figure}

The frequency-domain representation of a signal ($X_k$) carries information
about the signal's amplitude and phase at each frequency. Fig. \ref{fig:dft}
depicts the complex output information (power and phase). Notice that the
amplitude spectrum is the square root power spectrum. \\

\footnotetext{Notice that we convert the input image to gray-scale before applying FFT.}

\subsubsection{Azimuthal Average}
\begin{figure}[t]
\centering
   \includegraphics[width=\linewidth]{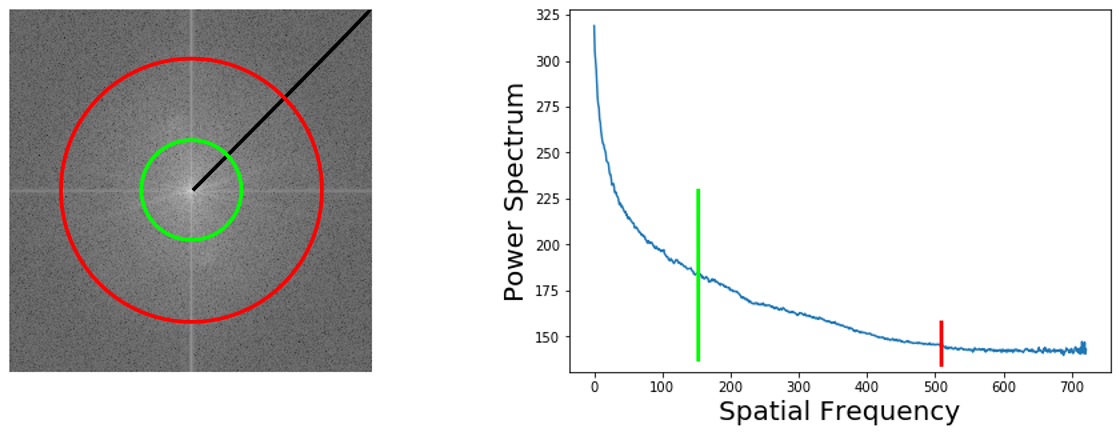}
   \caption{Example of an azimuthal average. (Left) Power Spectrum 2D. (Right) Power Spectrum 1D. Each frequency component is the radial average from the 2D spectrum.}
\label{fig:azi}
\end{figure}

After applying a Fourier Transform to a sample image, the information is represented
in a new domain but within the same dimensionality. Therefore, given that we
work with images, the output still contains 2D information. We apply azimuthal averaging
to compute a robust 1D representation of the FFT power spectrum. It can be seen as a
compression, gathering and averaging similar frequency
components into a vector of features. In this way, we can reduce the amount of features without
losing relevant information. Furthermore, throughout this compression, we
achieve a more robust representation of the input. Fig. \ref{fig:azi} shows a
visual example of such method.\\

\subsection{Classifier Algorithms}

Classification is the task to learn a general mapping from the attribute space to descrete classes, using specific examples of instances, each represented by a vector of attribute values and their acording lable. \\

\subsubsection{Logistic Regression}
One of the technically simplest (linear) classification algorithms is the Logistic Regression (LR). It is a simple statistical model that employs a logistic
function (see Formula \eqref{eq:1}) to model a binary dependent variable. The
output from the hypothesis $h$ is the estimated probability. This is used to
infer how confident predicted value can be given an input $\textbf{x}$.
Logistic regression is formulated as

\begin{align}
\begin{split}
        h_\textbf{w}(\textbf{x})= \dfrac{1}{1+e^{-\textbf{w}^T\textbf{x}}}
\end{split}
\label{eq:1}
\end{align}

The underlying algorithm of maximum likelihood estimation determines the
regression coefficient $\textbf{w}$ for the model that accurately predicts and
fits the probability of the binary dependent variable. The algorithm stops when
the convergence criterion is met or the maximum number of iterations is reached.\\

\subsubsection{Support Vector Machines}
Support Vector Machines (SVMs) \cite{boser1992training, cortes1995support} are
among the most widely used learning algorithms for (non-linear) data classification. The target
of the SVM formulation is to produce a model (based on the training data) which  will identify
an optimal separating hyperplane, maximizing the margin between different
classes. Given a training set of instance-label pairs
($x_i,y_i$), $i= 1,...,l$ where $x_i \in R^n$ and $\textbf{y} \in \{1, -1\}^l$,
Training of SVMs is implemented by the solution of the following optimization problem

\begin{align}
\begin{split}
	\min_{\textbf{w},b,\bm{\xi}} \qquad & \dfrac{1}{2}||\textbf{w}||^2 + C \sum\limits_{i = 1}^{l} {\xi_i}\\
	\mathrm{s.t. }\qquad & y_{i}(\textbf{w}^T\phi(x_i)+ b)\geq 1 - \xi_i, \\
	& \xi_i\geq 0,
\end{split}
\end{align}

where $\textbf{w}$ and $b$ are the parameters of our classifier, $\xi$ is the slack variable and $C >0$ the penalty parameter of the error term.

Here training vectors $x_i$ are mapped into a higher dimensional space by the function $\phi$. The training objective of SVMs is to find a linear separating hyperplane with the maximal margin in this higher dimensional space.\\

\subsubsection{K-Means Clustering}
While supervised classification algorithms like SVM and LR rely on labeled training example to learn a classification, we also want to test the detection performance in the absence of any labeled data.
Clustering is an unsupervised machine learning technique which finds
similarities in the data points and group similar data points together. The key
assumption is that nearby points in the feature space exhibit similar
qualities and they can be clustered together. Clustering can be done using
different techniques like K-means clustering.

The K-means objective function is defined as 

\begin{align}
\begin{split}
	J = \sum\limits_{k = 1}^{K} \sum\limits_{i = 1}^{m} || x_i - \mu_k||^2
\end{split}
\label{eq:1}
\end{align}

where $K$ and $m$ are the number of clusters and samples respectively.\\
A common approach to heuristically approximate solutions is to iteratively identify nearby features based on the
distances calculated from initial centroids $\mu$. Then, these features are assigned to the closest
cluster and the centroids are re-estimated. Since the amount of clusters is determined by the user, it can be easily employed in
classification where we divide data into $K$ clusters with $K$ equal to or greater
than the number of classes.

\section{Experiments}

In this section, we show results for a series of experiments evaluating the
effectiveness of our approach. First, we introduce a new high-resolution data set, called \emph{Faces-HQ}, together with
its training settings and experiments, and we discuss our results in detail.
In order to verify our approach, we also evaluate on the \textit{CelebA} data set \cite{liu2015deep}, which contains medium-resolution images, and on the \textit{FaceForensics++} data set \cite{roessler2019faceforensicspp}, which contains low-resolution video sequences.

\subsection{Faces-HQ}
\subsubsection{Data set}
to the best of our knowledge, currently no public data set is providing high resolution images with annotated 
fake and real faces. Therefore, we have created our own data set from established sources, called
\textit{Faces-HQ}\footnote{Faces-HQ data has a size of 19GB. Download: https://cutt.ly/6enDLYG}. In order to have a sufficient variety of faces, we have chosen to
download and label the images available from the \textit{CelebA-HQ} data set
\cite{karras2017progressive}, \textit{Flickr-Faces-HQ} data set \cite{karras2019style},
100K Faces project \cite{Faces}  and \textit{www.thispersondoesnotexist.com}. In total, we have
collected 40K high quality images, half of them real and the other half
fake faces. Table
\ref{tab:summary} contains a summary.\\

\begin{table}[h]
\centering
\resizebox{0.48\textwidth}{!}{\begin{tabular}{c|ccc}
\hline
& \# of samples & category & label \\
\hline
\rowcolor{Gray}
CelebA-HQ data set \cite{karras2017progressive} & 10000 & Fake & 0 \\
Flickr-Faces-HQ data set \cite{karras2019style} & 10000 & Fake & 0 \\
\rowcolor{Gray}
100K Faces project \cite{Faces} & 10000 & Real & 1 \\
www.thispersondoesnotexist.com & 10000 & Real & 1 \\
\hline
\end{tabular}}
	\caption{\textit{Faces-HQ} data set structure.}
\label{tab:summary}
\end{table}

\begin{figure}[t]
\centering
   \includegraphics[width=\linewidth]{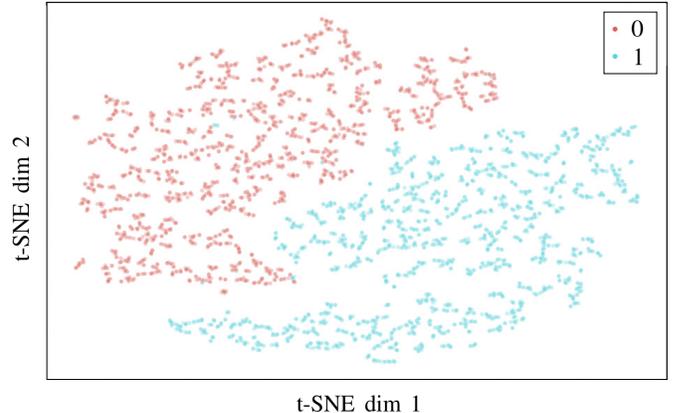}
	\caption{T-SNE visualization of 1D Power Spectrum on a random subset from \textit{Faces-HQ} data set. We used a perplexity of 4  and 4000 iterations to produce the plot.}
\label{fig:tsne}
\end{figure}

\subsubsection{Training Setting}

as shown in Fig. \ref{fig:pipeline}, our pipeline is split into two
parts. On the one hand, at pre-processing time, we take the whole data set and
we transform every sample from the spatial domain to the 1D frequency domain,
reducing 1024x1024x3 high quality color images to 722 features (1D Power
Spectrum). This method is formed by a Discrete Fourier Transform followed by an
azimuthally average. The transformation can be substantially optimized by
employing the Fast Fourier Transform. Notice that after applying the
transformation, we use only the power spectrum since it already contains enough
information for the classifier. 
A first visualization (see Fig. \ref{fig:tsne}) using t-Distributed Stochastic Neighbor
Embedding \cite{maaten2008visualizing} (t-SNE) reveals a clear clustering of fake and real samples in this feature space.

On the other hand, once the pre-processing step is finished, we start training
the classifier engine.  First of all, we divide  the transformed data into
training and testing sets, with  20\% for the testing stage and use the
remaining 80\% as the training set. Then, we train a classifier with the
training data and finally evaluate the accuracy on the testing set. Our goal is
to distinguish, real and fake faces, thus we need to use a binary
classifier.\\

\begin{figure*}

\begin{subfigure}{\linewidth}
\centering
  \includegraphics[width=\linewidth]{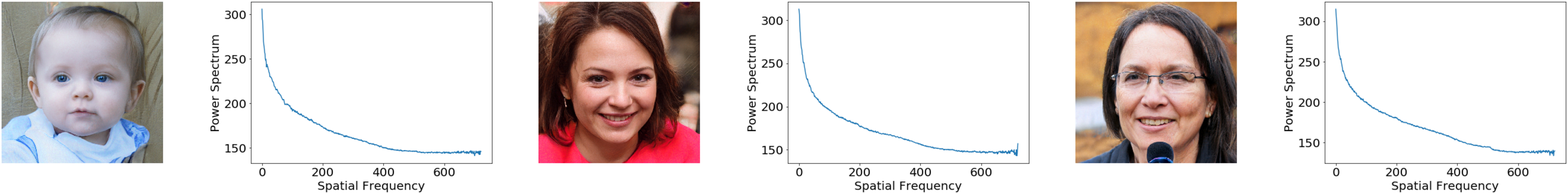}
  \caption{www.thispersondoesnotexist.com}\label{fig:1}
\end{subfigure}\\[2ex]

\begin{subfigure}{\linewidth}
\centering
  \includegraphics[width=\linewidth]{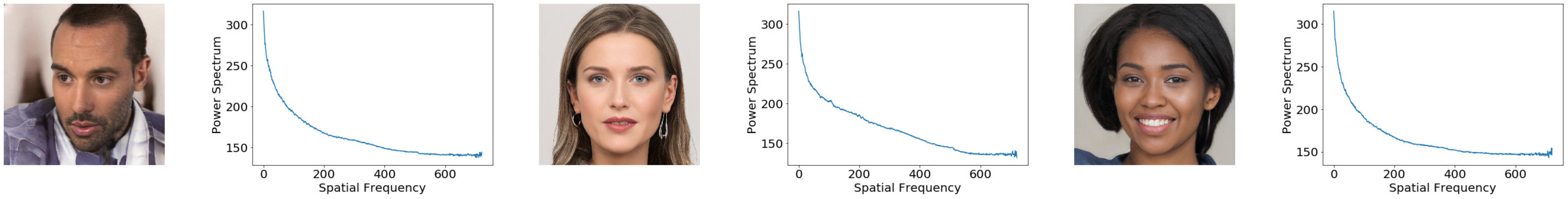}
  \caption{100K Faces project.}\label{fig:2}
\end{subfigure}\\[2ex]

\begin{subfigure}{\linewidth}
\centering
  \includegraphics[width=\linewidth]{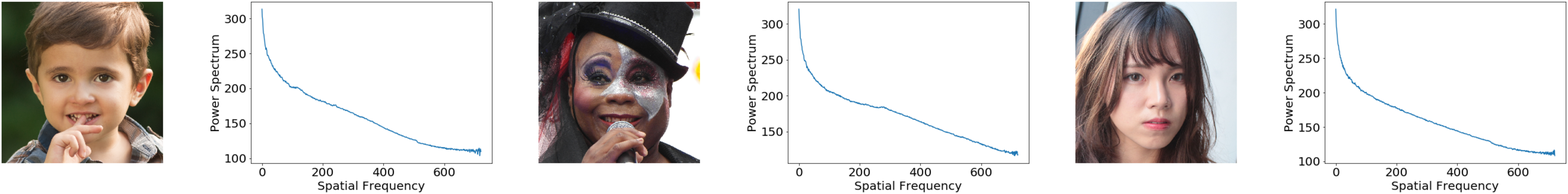}
  \caption{Flickr-Faces-HQ data set.}\label{fig:3}
\end{subfigure}\\[2ex]

\begin{subfigure}{\linewidth}
\centering
  \includegraphics[width=\linewidth]{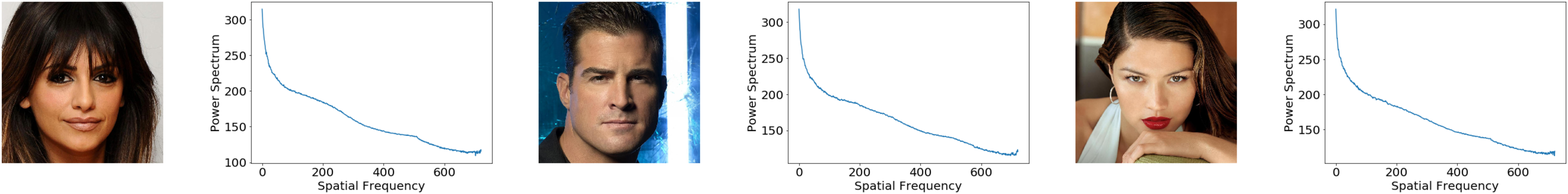}
  \caption{CelebA-HQ data set.}\label{fig:4}
\end{subfigure}

	\caption{Samples from  the different data sets gathered on \textit{Faces-HQ} data set. It is possible to observe on the 1D Power Spectrum some similitudes between images belonging to the same class and differences otherwise. For instance, real faces (\subref{fig:3}) and (\subref{fig:4}) do not have flat regions at high frequencies, whereas fake (\subref{fig:1}) and (\subref{fig:2}) have them.}
\label{fig:all}
\end{figure*}

\subsubsection{Method 1D Power Spectrum}

looking at Fig. \ref{fig:all}, one can observe that there is a certain
repetitive behavior or pattern on the 1D Power Spectrum on those images that
belong to the same class. Just by checking individual samples, it is possible
to conclude that real and fake images behave in noticeable different spectra at
high frequencies, and therefore they can be easily classified. 

\begin{table}[t]
\centering
\resizebox{0.35\textwidth}{!}{\begin{tabular}{c|ccc}
\hline
& \multicolumn{3}{c}{80\% (train) - 20\% (test)} \\
\cline{2-4}
\# samples & SVM & Logistic Reg. & K-Means  \\
\hline
\rowcolor{Gray}
4000 & 100\% & 100\% & 82\%  \\
1000 & 100\% & 100\% & 82\%  \\
\rowcolor{Gray}
100 & 100\% & 100\% & 81\%  \\
20 & 100\% & 100\% & 75\%  \\
\hline
\end{tabular}}
\caption{Test accuracy using SVM, logistic regression and k-means classifier under different data settings.}
\label{tab:both}
\end{table}

Driven by this phenomenon, we have evaluated a significant subset of images
(4000 in total, 1000 of each sub-data set) and we have computed basic statistics
to try to find a more general representation that help to simplify the problem.
Fig. \ref{fig:1000} plots the mean and the standard deviation of each
sub-data set and corroborates the observable and distinguishable trend that real
and fake images have.  Motivated by this  observations we have carried out a
set of tests to determine the extent to which our approach successfully
detects deepfakes and how much data is needed to train the model. In our experiments, we have
implemented one classifier based on support vector machines (SVMs) with a
radial basis function kernel and one based on logistic regression. We have run an
initial experiment using 80\% of the data for training and 20\% for testing. We
have utilized this configuration for different amount of samples (4000, 1000,
100, 20) equally distributed (see Table \ref{tab:both}).

\begin{table}[h]
\centering
\resizebox{0.48\textwidth}{!}{\begin{tabular}{c|ccccccc}
\hline
\backslashbox{from}{to}
& 100 & 200 & 300 & 400 & 500 & 600 & 722 \\
\hline
\rowcolor{Gray}
0 & 58\% & 69\% & 85\% & 89\% & 98\% & \textbf{100\%} & \textbf{100\%} \\
100 & - & 72\% & 86\% & 89\% & 98\% & \textbf{100\%} & \textbf{100\%} \\
\rowcolor{Gray}
200 & - & - & 85\% & 87\% & 99\% & \textbf{100\%} & \textbf{100\%} \\
300 & - & - & - & 84\% & 98\% & \textbf{100\%} & \textbf{100\%} \\
\rowcolor{Gray}
400 & - & - & - & - & 93\% & \textbf{100\%} & \textbf{100\%} \\
500 & - & - & - & - & - & \textbf{100\%} & \textbf{100\%} \\
\rowcolor{Gray}
600 & - & - & - & - & - & - & \textbf{100\%} \\
\hline
\end{tabular}}
\caption{Test accuracy using SVM classifier.}
\label{tab:svm}
\end{table}

\begin{table}[h]
\centering
\resizebox{0.48\textwidth}{!}{\begin{tabular}{c|ccccccc}
\hline
\backslashbox{from}{to}
& 100 & 200 & 300 & 400 & 500 & 600 & 722 \\
\hline
\rowcolor{Gray}
0 & 58\% & 70\% & 86\% & 90\% & 98\% & \textbf{100\%} & \textbf{100\%} \\
100 & - & 72\% &88\% & 90\% & 98\% & \textbf{100\%} & \textbf{100\%} \\
\rowcolor{Gray}
200 & - & - & 86\% & 89\% & 99\% & \textbf{100\%} & \textbf{100\%} \\
300 & - & - & - & 85\% & 98\% & \textbf{100\%} & \textbf{100\%} \\
\rowcolor{Gray}
400 & - & - & - & - & 92\% & \textbf{100\%} & \textbf{100\%} \\
500 & - & - & - & - & - & \textbf{100\%} & \textbf{100\%} \\
\rowcolor{Gray}
600 & - & - & - & - & - & - & 99\% \\
\hline
\end{tabular}}
\caption{Test accuracy using logistic regression classifier.}
\label{tab:log}
\end{table}

\begin{table}[h]
\centering
\resizebox{0.48\textwidth}{!}{\begin{tabular}{c|ccccccc}
\hline
\backslashbox{from}{to}
& 100 & 200 & 300 & 400 & 500 & 600 & 722 \\
\hline
\rowcolor{Gray}
0 & 37\% & 37\% & 55\% & 56\% & 62\% & 72\% & 82\% \\
100 & - & 39\% & 48\% & 57\% & 63\% & 72\% & 82\% \\
\rowcolor{Gray}
200 & - & - & 53\% & 61\% & 67\% & 73\% & 82\% \\
300 & - & - & - & 70\% & 72\% & 76\% & 85\% \\
\rowcolor{Gray}
400 & - & - & - & - & 75\% & 80\% & 89\% \\
500 & - & - & - & - & - & 83\% & 91\% \\
\rowcolor{Gray}
600 & - & - & - & - & - & - & \textbf{94}\% \\
\hline
\end{tabular}}
\caption{Test accuracy using k-means classifier.}
\label{tab:kmean}
\end{table}

After testing the effectiveness and efficiency of our transformed features, we
have conducted a another round of experiments to determine the impact of 
different frequency components.
Given the 722 features
from 1D Power Spectrum, we have analyzed the relevance of different frequencies by grouping them into
28 sub-sections. Table \ref{tab:svm},Table \ref{tab:log} and Table
\ref{tab:kmean} show the accuracy results on SVM, logistic regression and
K-means respectively. The rows indicate where the chunk of frequencies starts,
and the column where it ends. For example, there is a chunk with 0.86 accuracy
that contains frequencies from 100 to 300.

\subsection{CelebA}
\subsubsection{Data set}
CelebFaces Attributes (\textit{CelebA}) data set \cite{liu2015deep} consists of 202,599 celebrity face images with 40 variations in facial attributes. The dimensions of the face images are 178x218x3, which can be considered to be a medium resolution in our context.\\

\subsubsection{Training Setting}
In order to train our forgery detection classifier we need both real and fake
images. We used the real images from the \textit{CelebA} data set.
On the same set, we then train a DCGAN \cite{radford2015unsupervised} to
generate realistic but fake images. We
split the data set into 162,770 images for training and 39,829 for testing, and
we crop and resize the initial 178x218x3 size images to 128x128x3. Once the
model is trained, we can conduct the classification experiments on
medium-resolution scale.\\

\subsubsection{Results}

We follow the same procedure as in the previous experiments. Table \ref{tab:both3}) shows perfect classification accuracy in the supervised, and also very good results in unsupervised clustering.   
 
\begin{table}[h]
\centering
\resizebox{0.35\textwidth}{!}{\begin{tabular}{c|ccc}
\hline
& \multicolumn{3}{c}{80\% (train) - 20\% (test)} \\
\cline{2-4}
\# samples & SVM & Logistic Reg. & K-Means  \\
\hline
\rowcolor{Gray}
2000 & 100\% & 100\% & 96\%  \\
\hline
\end{tabular}}
\caption{Test accuracy using SVM, logistic regression and k-means classifier.}
\label{tab:both3}
\end{table}

\subsection{FaceForensics++}
\subsubsection{Data set}
\textit{FaceForensics++} \cite{roessler2019faceforensicspp} is a forensics data set
consisting of video sequences that have been modified with different automated
face manipulation methods. Additionally, it is hosting DeepFakeDetection
Data set. In particular, this data set contains 363 original sequences from 28
paid actors in 16 different scenes as well as over 3000 manipulated videos
using DeepFakes and their corresponding binary masks. All videos contain a
trackable, mostly frontal face without occlusions which enables automated
tampering methods to generate realistic forgeries.\\

\subsubsection{Training Setting}

The employed pipeline for this data set is the same as for \textit{Faces-HQ} data set and \textit{CelebA}, but with an additional block. Since the DeepFakeDetection data set contains videos,
we first need to extract the frame and then crop the inner faces from them. Due
to the different content of the scenes of the videos, these cropped faces have
different sizes. 

\begin{figure}

\begin{subfigure}{\linewidth}
\centering
  \includegraphics[width=\linewidth]{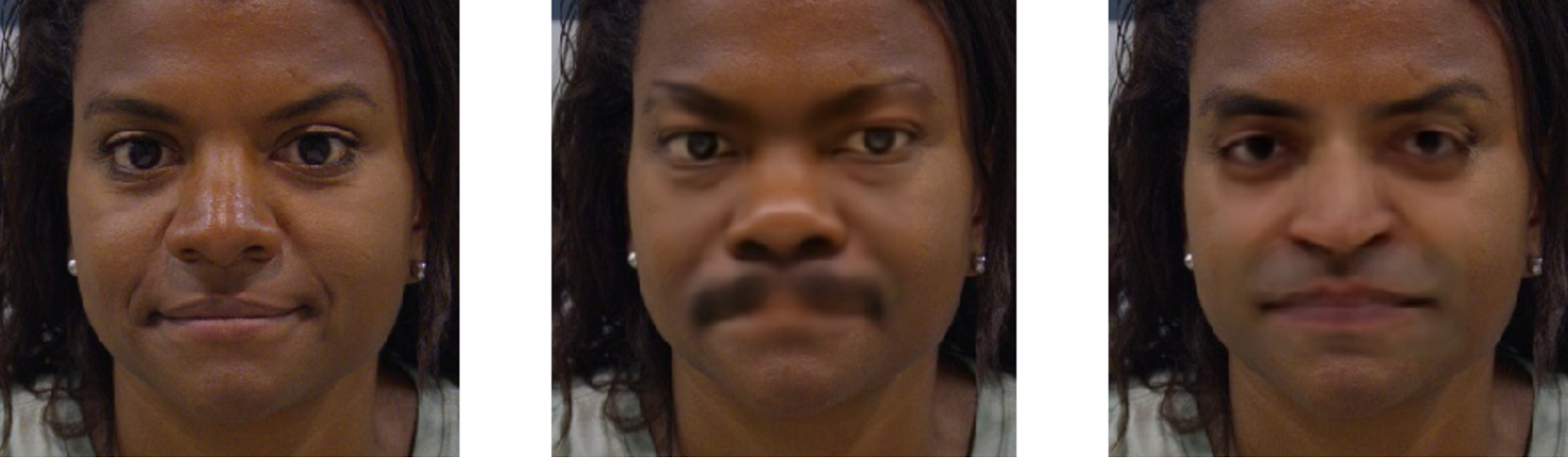}
  \caption{Example of one real face (left) and two deepfake faces, fake 1 (center) and fake 2 (right). Notice that the modifications only affect the inner face. }
  \label{fig:11}
\end{subfigure}\\[2ex]

\begin{subfigure}{\linewidth}
\centering
  \includegraphics[width=\linewidth]{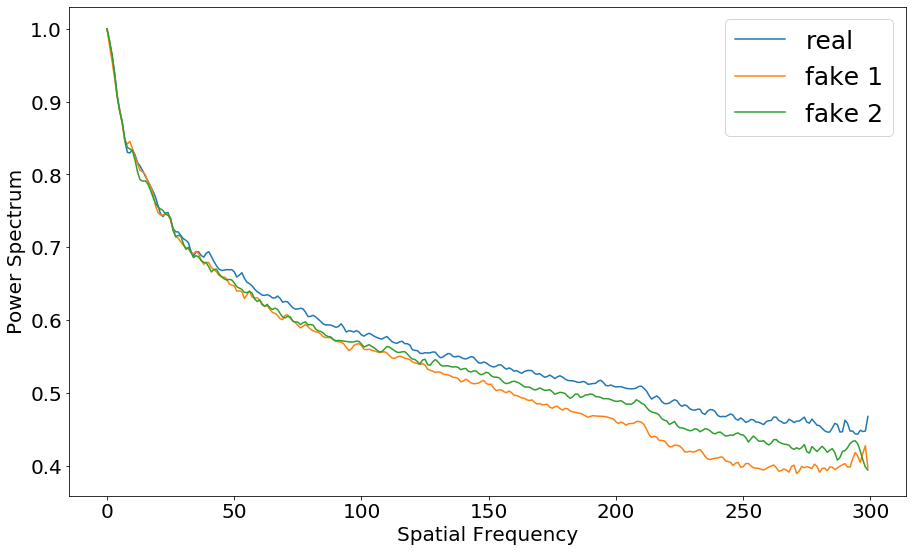}
  \caption{Normalized and interpolated 1D Power Spectrum from the previous images.}
  \label{fig:12}
\end{subfigure}\\[2ex]

\caption{Cropped samples from DeepFakeDetection data set and their corresponding statistics.}
\label{fig:deepfaces}
\end{figure}
The pre-processing part from the pipeline is size independent, thus no changes
are required. However, this is not true for the classifiers, since they expect a
fixed amount of features. Therefore, we have added an extra processing block
just before the classifier that interpolates the 1D Power Spectrum to a fix
size (300) and normalizes it dividing it by the $0$th frequency component.
The rest of the pipeline remains unchanged.\\

\subsubsection{Method 1D Power Spectrum}

As in the previous experiments, Fig. \ref{fig:deepfaces} shows that deepfake images have a
noticeably different frequency characteristic. Despite of having a similar
behaviour along the spatial frequency, there is a clear offset between the real
 the fakes that allows the images to be classified. 

\begin{table}[t]
\centering
\resizebox{0.3\textwidth}{!}{\begin{tabular}{c|cc}
\hline
& \multicolumn{2}{c}{80\% (train) - 20\% (test)} \\
\cline{2-3}
\# samples & SVM & Logistic Reg.\\
\hline
\rowcolor{Gray}
2000 & 85\% & 78\% \\
1000 & 82\% & 76\%  \\
\rowcolor{Gray}
200 & 77\% & 73\% \\
20 & 66\% & 76\% \\
\hline
\end{tabular}}
\caption{Test accuracy using SVM classifier and logistic regression classifier under different data settings.}
\label{tab:both2}
\end{table}

Table \ref{tab:both2}
contains the classification accuracy for the supervised algorithms. These
results confirm the robustness of frequency components as classification
features. Nevertheless, in this case, we have observed a slightly different
behaviour with respect to \textit{Faces-HQ} accuracy results (see Table \ref{tab:both}). 
The problem is become harder for low-resolution inputs. Hence,
the accuracy starts to decrease when the number of samples is smaller than
1000, specially, for the logistic regression. 

\begin{figure}[t]
\centering
   \includegraphics[width=\linewidth]{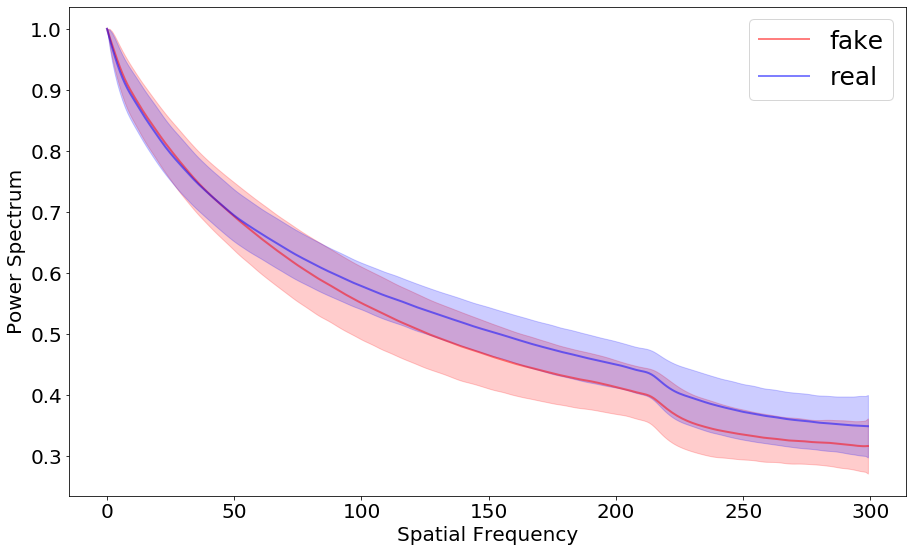}
   \caption{1D Power Spectrum statistics from DeepFakeDetection data set.}
\label{fig:2000}
\end{figure}

The dependency on samples and the non-perfect classification accuracy can be
understood by looking at Fig. \ref{fig:2000}. We can see how the standard
deviations from the real and the deepfake statistics overlap with each other, meaning that some samples will be
misclassified. As a result, it is not recommendable to reduce the number of
features, since now the classifiers are much more sensitive to the number of features.

Finally, we compute the average classification rate per video, applying a simple majority vote over the single frame classifications. 
Table \ref{tab:videos} shows the accuracy test results, which are relatively higher than 
the previous ones based on a frame by frame evaluation.\\

\begin{table}[h]
\centering
\resizebox{0.2\textwidth}{!}{\begin{tabular}{cc}
\hline
SVM & Logistic Reg.\\
\hline
\rowcolor{Gray}
90\% & 81\% \\

\hline
\end{tabular}}
\caption{Test accuracy per video using SVM classifier and logistic regression classifier.}
\label{tab:videos}
\end{table}

\section{Discussion and Conclusion}
In this paper, we described and evaluated the efficacy of a new method to
expose AI-generated fake faces images. Our approach is based on a
high-frequency component analysis. We
performed extensive experiments to demonstrate the robustness of our pipeline
independently of the source image. 
We show that our method is able to detect high- and medium-resolution deepfake images on two data sets with data from various GANs with 100\% accuracy. Low-resolution content is harder to identify since the available frequency spectrum is much smaller. Nevertheless, we are able to identify low-resolution fakes in a popular benchmark with 91\% accuracy. 

{\small
\bibliographystyle{ieee}
\bibliography{bibliography}
}

\end{document}